\documentclass{article}

\usepackage{microtype}
\usepackage{graphicx}

\usepackage{booktabs} 

\usepackage{amsmath}
\usepackage{amsfonts}
\usepackage{amssymb}
\usepackage{caption}
\usepackage{subcaption}

\usepackage{hyperref}

\usepackage[accepted]{icml2021}

\icmltitlerunning{Fixed $\beta$-VAE Encoding for Curious Exploration in Complex 3D Environments}

\begin{document}

\twocolumn[
\icmltitle{Fixed $\beta$-VAE Encoding for Curious Exploration in Sparse Reward 3D Environments}

\begin{icmlauthorlist}
\icmlauthor{Auguste Lehuger}{EP,ICL}
\icmlauthor{Matthew Crosby}{ICL}
\end{icmlauthorlist}

\icmlaffiliation{ICL}{Department of Computer Science, Imperial College, London, United Kingdom}
\icmlaffiliation{EP}{École polytechnique, Paris, France}

\icmlcorrespondingauthor{}{auguste.lehuger@polytechnique.edu}

\vskip 0.3in
]



\printAffiliationsAndNotice{}  

\begin{abstract}
Curiosity is a general method for augmenting an environment reward with an intrinsic reward, which encourages exploration and is especially useful in sparse reward settings. 
As curiosity is calculated using next state prediction error, the type of state encoding used has a large impact on performance. Random features and inverse-dynamics features are generally preferred over VAEs based on previous results from Atari and other mostly 2D environments. However, unlike VAEs, they may not encode \emph{sufficient} information for optimal behaviour, which becomes increasingly important as environments become more complex. In this paper, we use the sparse reward 3D physics environment Animal-AI, to demonstrate how a fixed $\beta$-VAE encoding can be used effectively with curiosity. We combine this with curriculum learning to solve the previously unsolved exploration intensive \emph{detour tasks} while achieving 22\% gain in sample efficiency on the training curriculum against the next best encoding. We also corroborate the results on Atari Breakout, with our custom encoding outperforming random features and inverse-dynamics features.
\end{abstract}

\section{Introduction}

Reinforcement learning relies on learning from a reward signal, which means that environments with sparse rewards can prove challenging. A domain general method to alleviate this problem is to augment the agent with an intrinsic reward signal called curiosity. Curiosity was first introduced in terms of an increase in knowledge of a network implementing a world model \cite{first}. Many related ideas have been studied, prominent is the recent formulation in terms of the error an agent receives when predicting the next state from current observations \cite{DeepakCuriosity}. From this self-supervised state prediction error, it was shown to be possible to solve certain problems without even including the predefined extrinsic reward of the environment. Since then, curiosity of some form has been used in many state-of-the-art algorithms across a wide variety of domains, (e.g.\ DeepMind's Agent 57 amongst many others) \citep{agent57,huang2019learning,zhao2019curiosity,luo2019curiosity}. Nowadays, many RL libraries include curiosity as a default setting because it often improves sample efficiency allowing RL to scale to more complex problems \cite{juliani2018unity}. 

Curiosity rewards are typically given for entering either novel  \cite{bellemare2016unifying} or surprising states \cite{achiam2017surprise, DeepakCuriosity}. We focus on the latter here, especially the case where surprise is defined in terms of the next state prediction error. This makes the metric particularly dependent on the state encoding. Previous work has shown that ideally the encoding should be \emph{compact} (reduce dimensionality), \emph{sufficient} (contain all necessary information), and \emph{stable} (not change over time) \cite{burda2018exploration}. Using raw pixels instead of an encoding in not \emph{compact}. There might be a large unpredictable change in the environment that is not relevant for agent behaviour, yet spikes the curiosity reward. Using random convolutions is more \emph{compact}, but may not be \emph{sufficient}. A promising idea to have both \emph{compactness} and \emph{sufficiency} is to use $\beta$-VAE encodings, but this runs into problems of \emph{stability}. 

As an agent improves its policy it will naturally start to explore different areas of the state space. A fixed VAE trained over an agent's initial observations will quickly become out of date. On the other hand, an online VAE that is continually changing as the agent explores leads to an extremely volatile reward signal, which hinders learning, and even exploration. This has meant that in domains such as Atari, random features are used instead, as they have been empirically shown to be more effective even though they may not be \emph{sufficient}. We hypothesise that for more complex domains, sufficiency will become more important. In simple domains, adding even random exploration is a known strategy that can improve algorithm performance, with methods such as epsilon-greedy implementing this directly. But, for more complex domains, a sufficient encoding has the potential to allow for more directed exploration, so that the state space is explored more methodically.

Humans appear to come with `developmental start-up software' which provides foundational knowledge of core domains early in development \cite{lake2017building}. Many animals are born with innate knowledge that allows them to navigate their environment, sometimes in full control of their bodies within minutes of birth. Chicks only a few days old can solve object permanence and spatial elimination tasks, seemingly inferring the location of an object that they have been imprinted upon that has moved out of sight \cite{chiandetti2011intuitive}. As animals explore the particular environment they have found themselves in they also create an adaptive encoding. For example, the chicks in the experiment above have adapted to imprint on objects used for testing instead of their mothers. They use both a more stable innate knowledge that contains information general to the evolution of the species (e.g.\ intuitive physics on Earth) and an adaptive representation specific to the individual. Our experiments suggest that the innate encoding may do more than just give a headstart to learning. It may also provide a \emph{stable} foundation for curious behaviour.

Innate knowledge can be thought of as the stable encoding of a pretrained VAE that builds general principles into the system. This VAE should be trained on domain general knowledge, such as random samples from a pre-defined open world environment before any task specific details are included. In this paper we make the first step of showing that this fixed encoding can be used with curiosity to improve performance in animal-inspired navigation tasks. Ultimately, in the future we hope that this stable innate knowledge, which represents general encoded knowledge through evolution can be augmented with unstable adaptive representations that are agent specific.

In this paper we work with exploration intensive tasks in a 3D-physics environment, Animal-AI,  and show that, where it is possible to create a fixed VAE encoding that remains representative as the agent improves, this can be used with curiosity to increase performance. This allows us to solve previously unsolved tasks, where random features are not sufficient. We use Animal-AI for our experiments as it was developed to create a sandbox for testing agents in a 3D environment specifically geared towards cognitive tasks \cite{AnimalAI-Testbed, shanahan2020artificial}. Whilst this is still a large oversimplification when compared to the real world, we hypothesise that it is the kind of environment and contains the kind of tasks where curiosity can play a stronger role. Animal-AI was also designed to ensure a large transfer gap between training and test tasks, originally released with only held-out tasks. In particular we work with \emph{detour tasks}, in which an agent must navigate around a (possibly previously unseen) obstacle in an efficient manner. Such tasks have been used to test many animals and are used to measure spatial problem solving \citep{smith2010well,maclean2014evolution}.

Our main contribution is to show the benefits of using curiosity with an offline trained $\beta$-VAE. This VAE encoding is compared to fine-tuned IDF and RF encoders in a sparse reward exploration-intensive environment. In this setting, we demonstrate that the fixed $\beta$-VAE encoding scores better than RF and IDF encoding on three key metrics: \textit{sample-efficiency}, \textit{training score} and \textit{test score}. The $\beta$-VAE required, on average, 22\% less training steps than the fine-tuned baseline obtained with state-of-the-art curiosity module. It also scored better than the baseline on training environments as well as unseen environments. We also corroborated these results on Atari Breakout. Whilst this environment is not ideal for training an offline VAE, we confirm that if one was available, then curiosity can exploit its inherent structure to achieve better performance than using random features or inverse dynamics features. 

The rest of the paper is structured as follows. We first survey the related literature, and in particular the introduction of the three criteria mentioned above for evaluating encodings for use with curiosity. In Section \ref{sec:background} we introduce the relevant technical background for the rest of the paper as well as the types of tests we are using. Section \ref{sec:experiments} outlines the experimental setup including the curriculum design and the VAE pre-training. We then report and analyse the results, before suggesting future work and concluding the paper.

\section{Related Work}\label{sec:relatedwork}

In this section we survey the related work on curiosity, VAEs, and curriculum learning, the three main components of our work. As the combined literature is vast, we only mention the most relevant papers here.

The ideas behind curiosity have been around for a while as it speaks directly to the exploration problem for training intelligent agents \cite{first}. There are also many forms of curiosity, all different methods of augmenting an agent's extrinsic reward function to encourage exploration. This can be done directly, by keeping counts of each state that has been previously visited \cite{bellemare2016unifying}, or indirectly, by encouraging the exploration of surprising states \cite{achiam2017surprise, DeepakCuriosity}. The latter is the case we focus on here. It is perhaps more biologically plausible, and is more closely tied to work on the  minimisation of prediction error or surprise in neuroscience \cite{friston2009free}. 

Burda et al.\ performed a large-scale study of curiosity driven learning over 54 standard benchmarks in RL \cite{CuriosityReview}. An important contribution was the comparison of four possible encoding functions:\par
\textbf{Pixels}: the visual observations are used directly.\\
\textbf{Inverse Dynamics Features (IDF)}: the encoding network is learned by back-propagating the error of a subsequent inverse dynamics network. This network's task is to predict the action taken given the previous and current state.\\
\textbf{Random Features (RF)}: the embedding consists of a convolutional network that is fixed after random initialisation.\\
\textbf{Online Variational Autoencoder (VAE)}: a variational inference network  is trained in an auto-supervised manner and the learned encoding is then used as the embedding network \citep{kingma}.

The authors also introduced three ways of characterising the properties of encodings:\par
\textbf{Compactness}: Features should be low-dimensional.\\
\textbf{Sufficiency}: Features should contain all relevant information to solve the problem.\\
\textbf{Stability}: Features should remain stable across time.

Across their benchmarks it was found that pixels performed worse than the more \emph{compact} alternatives. This is mostly due to the forward model’s inability to predict the next pixel frame from the last frame and the action taken. VAE was next worst, being too unstable to generate a reliable signal. RF and IDF performed best, with the conclusion that RF is a ``surprisingly strong baseline''. 

The four types of encoding above are only a small sample of possible encodings. For example, \cite{anand2020unsupervised} introduce a method that learns state representations by maximizing mutual information across spatially and temporally distinct features of a neural encoder of the observations. In there experiments they also find that random features are surprisingly effective in Atari games.

It has become somewhat common to use VAEs to encode high-dimensional state spaces for use with DRL \cite{ha2018world}. 
Other work has explored the \emph{sufficiency} improvement VAE can bring when combined with curiosity. For example, Han et al.\ show how an online VAE can be combined with curiosity and augmented to DQN on Atari \cite{han2020curiosity}. They argue that curiosity is especially useful to guide the agent to sample new states for the VAE. This is compared to our work where we explore the use of a pre-trained $\beta$-VAE that remains fixed for use with curiosity.

Similar to our multi-stage approach, Higgins et al.\ created an agent (DARLA) to explicitly learn disentangled representations for environments where it is possible to train representations before training the agent \cite{pmlr-v70-higgins17a}. Their agent first learns features using a $\beta$-VAE and then uses them across multiple environments. Their main experiment uses DeepMind Lab \cite{beattie2016deepmind} which, like Animal-AI, has the useful feature of being able to specify an environment, from which observations can be drawn, before specifying the configurations of the environment to be used for testing. 
They show that the factorised representation learned by the $\beta$-VAE discovers can be used to encode general `factors of variation' of the environment and show that this, combined with its \emph{compactness} and pre-training method, allows them to capture features that are not task or domain specific allowing better zero-shot generalisation. We build on this result for our work combining $\beta$-VAEs with curiosity, with the idea that such a factorisation will lead to a curiosity signal that is more responsive to the relevant features of the environment.

Curriculum Learning (CL) is a general approach to optimising the learning process of a machine learning algorithm \cite{curriculum-learning}. In practice, it involves training on a sequence of tasks such that the ordering of tasks is designed to guide the learning progress. Justesen et al.\ demonstrated that curriculum learning can improve generalisation and avoid overfitting to a specific problem \cite{curriculum-difficulty}. In curriculum learning a common problem is catastrophic forgetting, where agents unlearn earlier capabilities. One solution is to interlace a curriculum with random tasks \cite{curriculum-forget}. In this paper we cycled through a curriculum multiple times and found that the resulting agent performed robustly across the full curriculum.

\section{Background}\label{sec:background}

A Deep Reinforcement Learning problem is usually defined over a Markov Decision Process (MDP), as a tuple $\langle S, A, P, R, \gamma \rangle$ where $S$ is the state space, $A$ is the set of actions, $P: S \times A \times S \rightarrow \mathbb{R}$ specifies the transition probabilities, $R: S \times A \times S \rightarrow \mathbb{R}$ is the reward function, and $\gamma \in [0, 1]$ is the discount factor \cite{sutton1998introduction}.

We will use $s_t$, $a_t$, $r_t$ to denote respectively the original state, the action chosen and the reward obtained at time t. Action $a_t$ is chosen by sampling from the policy network distribution: $\pi_\theta$. Trajectory $\tau$ = ($s_0$, ..., $s_T$) are the successive states reached when following the policy. $\pi_\theta(\tau)$ denotes the probability to end up with trajectory $\tau$ from $\pi_\theta$ and $r(\tau)$ denotes the discounted global rewards (weighted sum of intrinsic and extrinsic signal) obtained during the trajectory $\tau$.

Policy gradient algorithms aim at optimising a policy network, parameterised by $\theta$, through the maximisation of the objective function: $$J(\theta) :=  E_{\tau \sim \pi_{\theta}} [R^{\tau}] = \int_{\tau} \pi_{\theta}(\tau) r(\tau) d\tau$$  
The idea is to optimise the expectation of the cumulative rewards variable $R$, i.e.\ get the highest rewards $r$ from trajectories $\tau$ drawn from policy $\pi_{\theta}$. We will focus here on on-policy learning, where the agent explores by sampling trajectories $\tau$ from the latest version of its policy $\pi_{\theta_{old}}$.

Proximal Policy Optimisation (PPO) is considered to be a state-of-the-art policy gradient algorithm and is widely used in RL as it is easy to implement, does not require significant hyperparameter tuning and has good empirical support \cite{PPO}. The key idea behind PPO is to constrain the absolute relative difference between two consecutive policy updates to a factor $\epsilon$. We use PPO for all experiments in this paper.

Variational Auto-Encoders (VAE) are generative models capable of learning unsupervised latent representations of complex high-dimensional data \cite{kingma}. The VAE model consists of two symmetrical networks: the encoder $q_\phi(z|x)$ downsamples a high-dimensional input into multiple feature maps while the decoder $p_\theta(x|z)$ upsamples a set of features maps back to a high-dimensional output. The VAE objective is to learn the marginal likelihood of a sample x from a gaussian distribution parameterised by generated vector $z$. 
Here, we consider $q_\phi(z|x) $, the
variational approximation of $p_\theta(x|z)$, the true posterior. The marginal likelihood can be written as :
$\textrm{log } p_\theta(x) = \mathcal{L}(x; \theta, \phi) + D_{KL}(q_\phi(z|x)||p_\theta(z|x))$

Since the true data likelihood is usually intractable, instead, the VAE optimizes an evidence lower bound (ELBO) which is a valid lower bound of the true data log likelihood, defined as:

$ \mathcal{L}(x; \theta, \phi) = \mathbb{E}_{q_\phi} [\textrm{log } p_\theta(x|z)] - D_{KL}(q_\phi(z|x)||p(z)) $

The first term can be considered as reconstruction loss, and the second term is approximated posterior $q_\phi(z|x)$ from prior $p(z)$ via KL-divergence. A $\beta$-VAE is a variation of a VAE which adds a weight $\beta > 1$ to the KL divergence minimisation objective, which has a critical impact on the latent space shape \cite{beta-vae}. In 2017, this algorithm outperformed state-of-the-art algorithm in learning a disentangled factorised latent space on a variety of datasets with respect to most metrics.

In the AAI testbed, like most environments, rewards are sparse:  the agent receives an extrinsic reward only when it encounters a red, gold or green goal object. Therefore, it seems critical to augment the reward signal the agent receives to guide it through its training. In this paper, curiosity rewards are computed by an Intrinsic Curiosity Module (ICM), structured in two parts: an Encoding Network and a Forward Model \cite{DeepakCuriosity}. The former is a $\beta$-VAE encoding network that transforms the pixel-based state $s_t$ into a vector-based state $\phi(s_t)$, while the latter takes the current state encoding $\phi(s_t)$ and $a_t$ as inputs and predicts the feature representation $\widehat{\phi(s_{t+1})}$ with a a series of feed forward layers. The curiosity reward is defined as: 
$r_t^i = \|\widehat{\phi(s_{t+1})} - \phi(s_{t+1})\|$

\subsection{Detour Tasks}
Having covered some of the technical background we now briefly introduce detour tasks, and explain why we chose them to evaluate the use of VAEs with curiosity. The objective of a detour task is to test the ability of the agent to make a detour around an object via the shortest path. A common variation is the cylinder task, where the object is transparent, so that the agent must move away from a visible goal in order to actually reach it. 

MacLean et al.\ performed a large scale study of cylinder tasks on animals combining results of many empirical studies \cite{maclean2014evolution}. In particular, they characterised the problem as one of self control. The animal must \emph{inhibit} the natural response to go towards the rewarding object so that they can go around the barrier and eventually reach it. We focus on X-Maze Detour Tasks, where the animal (or agent) is placed in one quadrant of an X, and the goal in another quadrant \cite{smith2010well}. The length of the arms determine the difficulty of the task, with long arms requiring the agent to traverse a long way from its objective in order to actually reach it.

Detour tasks have been recreated in the Animal-AI testbed for use with AI systems. They were part of the 2019 competition where the maximum score was $12$ out of $36$ tasks solved and average top-10 score was $12\%$ \cite{AnimalAI-Testbed}.

\section{Experimental Setup}\label{sec:experiments}

We now describe the experimental setup for testing curiosity on detour tasks. First, we explain the design of the curriculum, followed by the training of the $\beta$-VAE itself.

\subsection{Curriculum Learning}

\begin{figure*}[t]
    \begin{subfigure}{.49\textwidth}
        \centering
        \includegraphics[width=\textwidth]{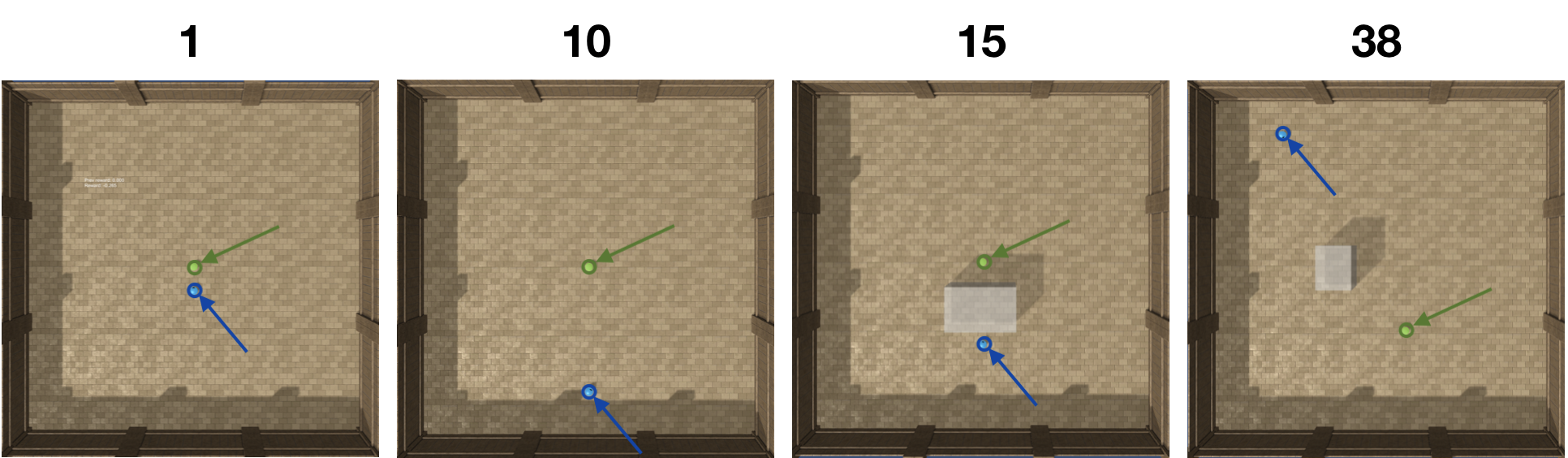}
        \caption{IDC}
        \label{fig:idc}
    \end{subfigure}
    \hfill
    \begin{subfigure}{.49\textwidth}
        \centering
        \includegraphics[width=\textwidth]{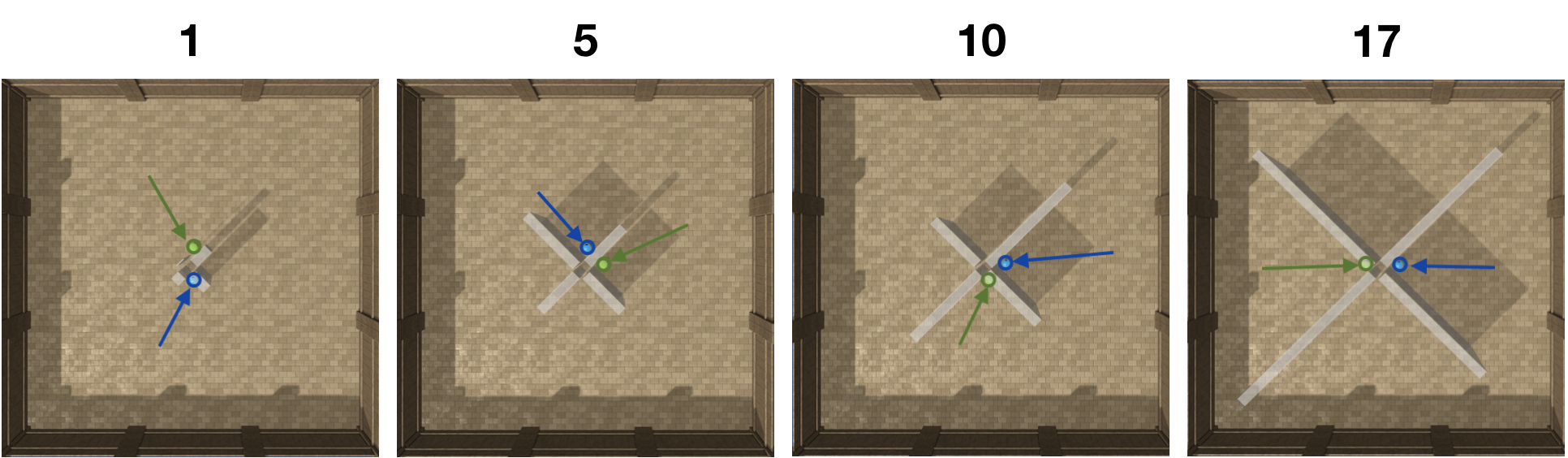}
        \caption{XMC}
        \label{fig:xmc}
    \end{subfigure}
    \caption{Representative examples from the two curriculum used to train the agents. These are AAI arenas seen from above. The agent is a small blue circle and is rewarded only upon reaching the food (small green circle).}
    \label{fig:curriculum}
\end{figure*}

As we are interested in testing how well curiosity works with an encoding that represents innate knowledge about the salient features of an environment, we decided to use a curriculum of increasing difficulty for the agents so that their behaviour can slowly adapt to the complexity of the tasks. The idea is that if we just used randomly created arenas for training (or train directly on the test set) then it is unlikely for curiosity to be as useful as prediction error is more common due to larger variance in the inputs. On the other hand, if we slowly increment tasks, then the forward model can `keep up' with the changes in inputs and prediction error is more likely to encode task relevant information. Additionally, we performed initial experiments without a curriculum and did not manage to train an agent to solve the tasks directly.   

We used two different curricula during the experiments. Figures ~\ref{fig:idc} and ~\ref{fig:xmc} show sample problems from both. The first, the Initial Detour Curriculum (IDC) contains 40 configurations designed to teach an agent basic abilities such as go towards food, reduce speed to turn, rotate towards food and avoid walls. These are not meant to be challenging problems. In fact, we found that using no curiosity is more effective than even  a small weighting to curiosity on the IDC. This is presumably because of the simplicity of the tasks and incremental nature of the curriculum such that exploitative behaviour is preferred (full details of these experiments are in the supplementary material). 

The IDC was used as a pre-training phase for each agent prior to involving curiosity and was identical for all the agents in the experiments. This meant the starting point for the experiments could be an already minimally competent agent so that we can test the use of curiosity with some innate knowledge and so that the prediction error is more directed towards environment specific correlations. There was little variance between six trained agents, with all completing the IDC in around 800K environment steps with similar learning curves.

The agents are trained using the PPO algorithm. The policy model comes in three parts. First, a series of CNNs downsamples the visual observation to a compact vector, which is then concatenated to the output of an LSTM recurrent network, fed with sequences of length 64.  Finally, the resulting vector goes through a single dense layer outputting action logits. The network is trained using Adam optimizer with a linearly decreasing learning rate, initialized at 3e-4. The epsilon clipping ratio is set at 0.2 and the gamma discount factor at 0.99. During the IDC training, agents are trained with a curiosity weight of 0. The agent moves to the next task in the curriculum once they achieve an extrinsic reward above a difficulty-dependent threshold value and the full curriculum is cycled through twice. 

The second curriculum is the X-Maze Curriculum (XMC), and is made of 18 variations on X-Maze tasks of increasing difficulty as shown in Figure ~\ref{fig:xmc}. The size of the maze increase between tasks, and the agent's starting position, rotation and exact length of the arms vary within multiple episodes of a given task. For the XMC training we use a curiosity weight of 0.01 and cycled the curriculum as many times as possible within 2.5 million steps. 

\subsection{Training $\beta$-Variational Auto-Encoder}

\begin{figure}[t]
\centering
\includegraphics[width = 1\hsize]{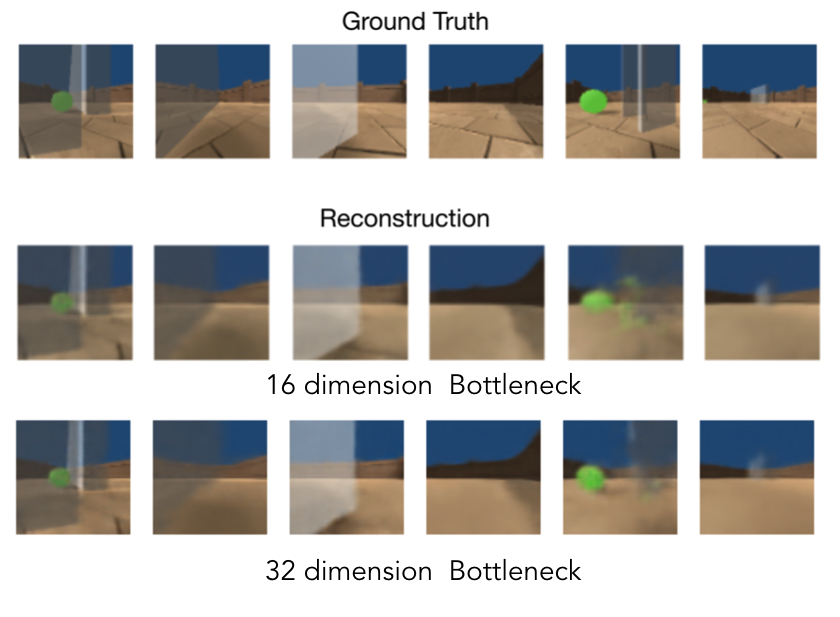}
\caption{Reconstruction of 6 randomly sampled images with differently sized latent space.}
\label{fig:vae-reconstruction}
\end{figure}

\begin{figure*}[t]
\begin{subfigure}{.49\textwidth}
    \centering
    \includegraphics[width=.7\textwidth]{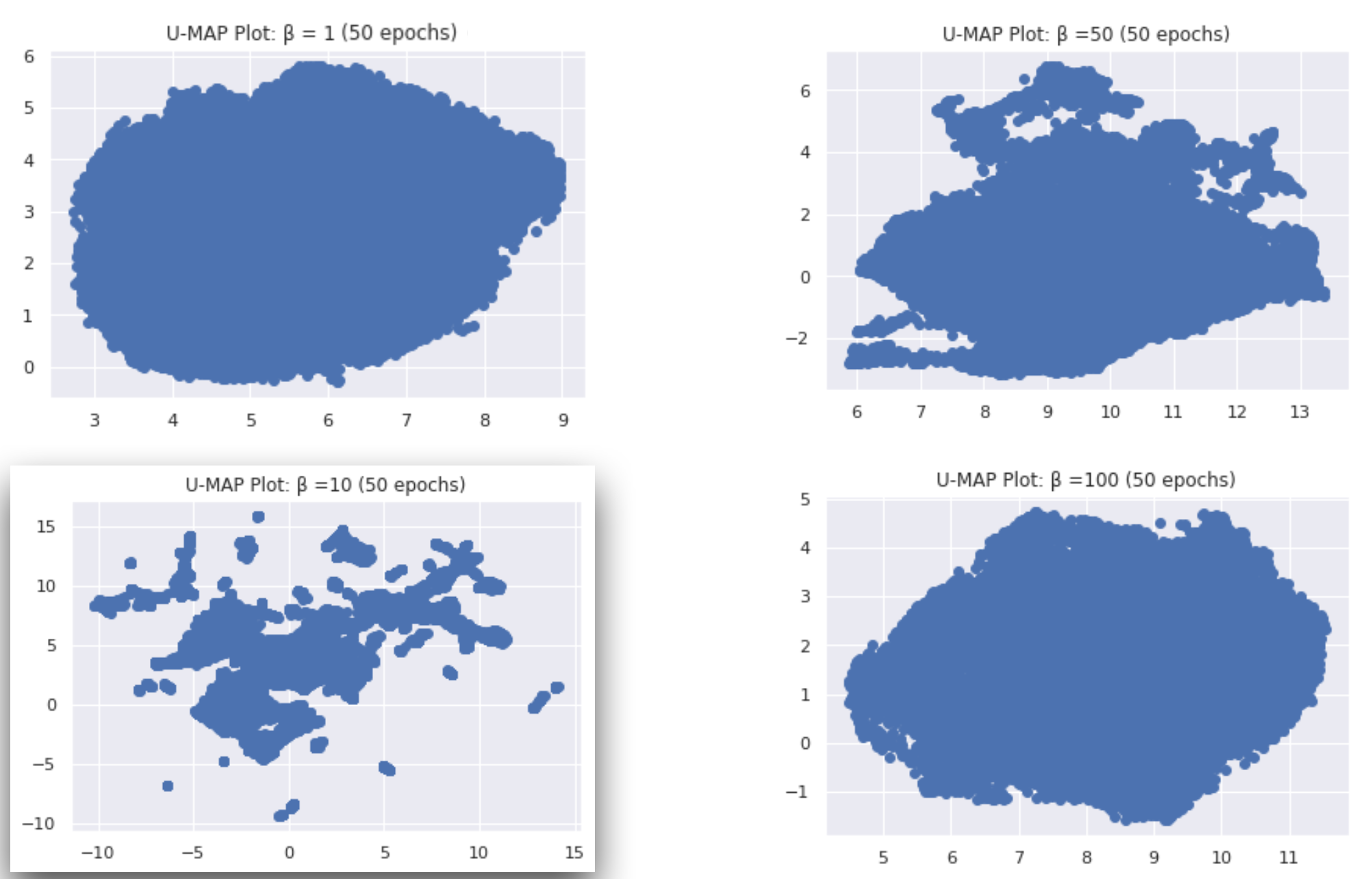}
    \caption{UMAP plots for varying $\beta$ values.}
    \label{fig:u-maps}
\end{subfigure}
\begin{subfigure}{.49\textwidth}
\centering
\includegraphics[width=\textwidth]{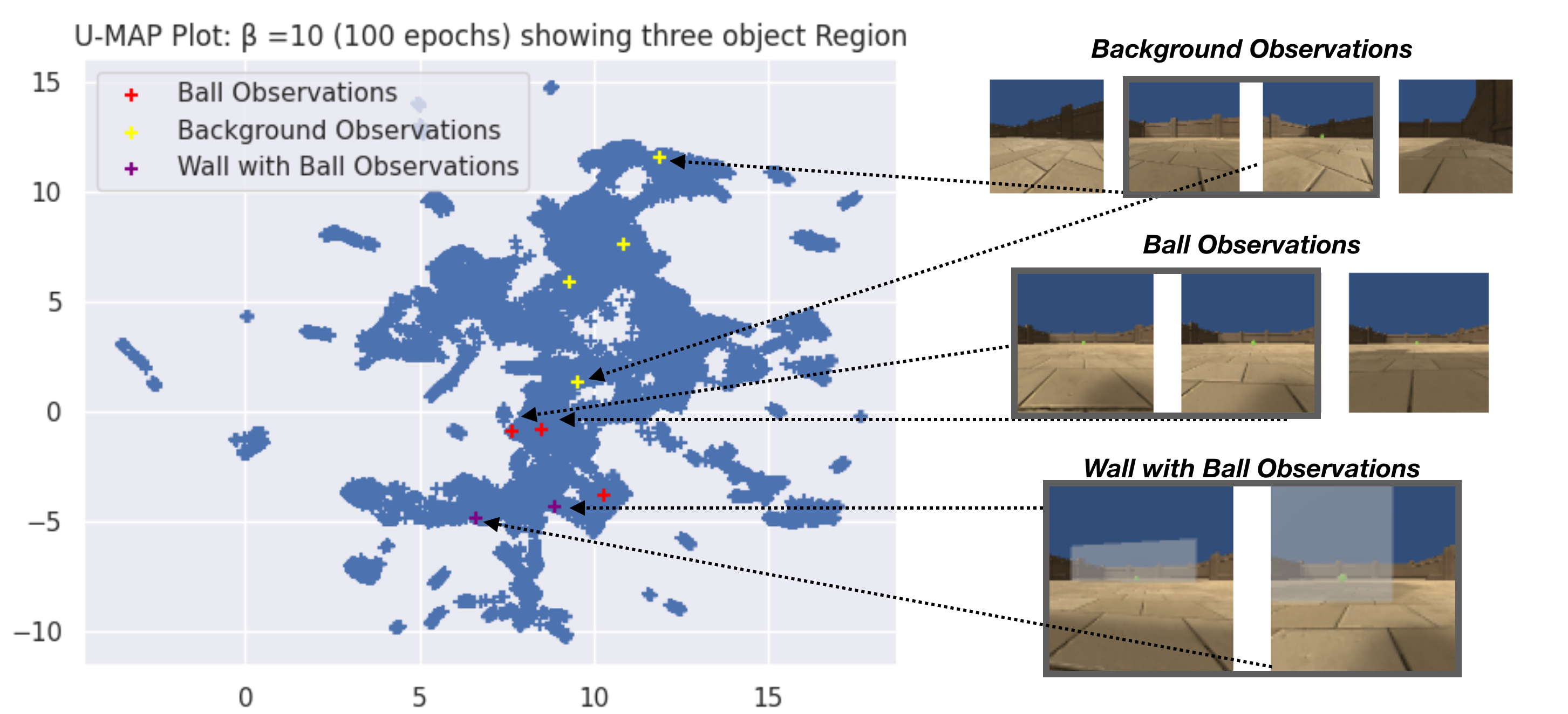}
\caption{UMAP of the chosen encoding.}
\label{fig:factorized-representation}
\end{subfigure}
\caption{Visualisations of the feature space of the $\beta$-VAE encodings. (a) shows UMAP plots for $\beta$ values $1, 10, 50, 100$. (b) shows how similar observations are grouped by the encoding.}
\end{figure*}

The benefit of using Animal-AI is that it is easy to generate a representative dataset of observations before training any agents. All experiments take place in the same static arena and the starting configuration can easily be chosen to randomise the agent position. This means that it is possible to sample from all states that the agent could see, even before it is competent enough to reach them. Compare this to most Atari games where an agent that is acting randomly only sees that beginning of the game repeatedly before dying. The data collection and pre-training step on Animal-AI requires negligible compute compared to the training of the DRL agents. The top row of Figure ~\ref{fig:vae-reconstruction} shows randomly selected observations from the generated dataset. The full set used included $60,000$ $84 \times 84 \times 3$ RGB images. 

We used a 4 layer convolutional network for the VAE encoding architecture. The latent dimension affects both the \emph{compactness} and \emph{sufficiency} of the encoding. Set too high, the encoding is not compact. Set too low, it is not sufficient. We assess the sufficiency qualitatively by observing if the VAE can fully reconstruct an image. We performed that latent dimension optimisation using this above $\beta$-VAE network trained for 100 epochs with $\beta = 10$. Figure ~\ref{fig:vae-reconstruction} shows the reconstruction of sample images for a latent dimension of 32 compared to at 16. We set the dimensionality at 32 for the experiments.

To explore the structure of the representations we also trained four $\beta$-VAE for 50 epochs with different $\beta$ factors: 1, 10, 50, 100 and assessed their factorisation. We used Uniform Manifold Approximation and Projection (UMAP) to project the data onto a Riemannian manifold of dimension two, where a distance metric can be applied \cite{2018arXivUMAP}. Figure ~\ref{fig:u-maps} shows the UMAP visualisation of all the dataset observations after being encoded by a different $\beta$ network. We observe that a $\beta$-VAE with $\beta = 10$ gives the best latent space factorisation. 

Ideally, the VAE will encode images that are relevantly similar for task performance close together. This way, large changes in the encoding space will represent large task-relevant changes in the raw inputs and, if not correctly predicted, lead to high curiosity values. On the other hand, common observations or extraneous details changing should give a low change in the coding space and therefore a low curiosity value even when not predicted perfectly. 
As there are three main objects in our environments (goal, outer walls, transparent walls) we analysed the factorisation to see if similar images in terms of clusters were factored together. Figure ~\ref{fig:factorized-representation} shows a UMAP visualisation annotated with images from the environment. The latent space qualitatively shows some semantic separation between the categories. 

With the setup of a VAE-encoding that has been fine-tuned on the state space and an agent pre-trained on the IDC to have basic non-Xmaze solving performance, we now have our fixed innate knowledge proxy in place and can analyse the impact of a more structured encoding with curiosity on a physical reasoning task.

\section{Animal-AI Results}\label{sec:resultsAAI}

\begin{figure}[t]
\centering
\includegraphics[width=.5\textwidth]{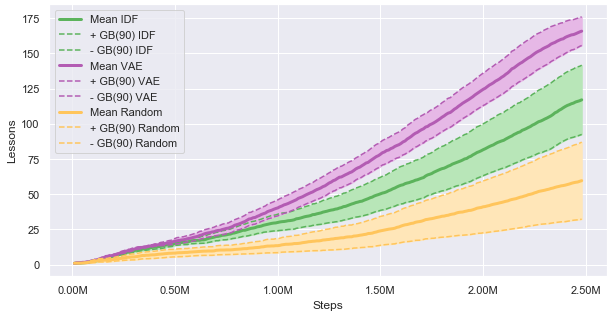}
\caption{Mean Confidence Interval 90\% for VAE, IDF, and RF agents learning the X-Maze curriculum.}
\label{fig:exp6_mean}
\end{figure}

We now present the results of our experiments on Animal-AI. We compare our agent to two baseline agents, IDF, and RF. We examine sample-efficiency, training score and test score in order to give a complete picture of the performance. 

\subsection{Sample-Efficiency}

Using a non-curious agent trained on the IDC as a starting point, we then trained each agent with curiosity on the XMC for 2.5M steps. Figure \ref{fig:exp6_mean} summarises the results. It can be seen from the graph that the VAE agent outperformed both the agent using IDF and Random Features over the training curriculum. We observe a gain of 22.8\% sample efficiency of VAE against IDF and 49.2\% of VAE against RF. Furthermore, the $\beta$-VAE method has the lowest variance with a standard deviation of 19.57 against 47.24 for IDF and 52.64 for Random. 

These results highlight the potential of the fixed $\beta$-VAE encoding for computing curiosity in these environments. We also note that the $\beta$-VAE had the lowest variance amongst the approaches suggesting that its stability is working as intended.

\subsection{Training Score}

\begin{table}[t]
    \caption{Performance on training and testing arenas with standard deviation.}
    \label{tab:AAIresults}
    \large
    \centering
        \begin{tabular}{rcl}
             & Training Score & Test Score \\
        \hline
        VAE & $93.6\% \pm 0.046$ & $46.2\% \pm 0.088$\\
        IDF & $87.1\% \pm 0.113$ & $38.2\% \pm 0.06$\\
        RF & $55.9\% \pm 0.334$ &  $29.1\% \pm 0.178$\\
        \end{tabular}
\end{table}

The previous results show how efficiently the agent moved through the curriculum during training. We also report the final success rate over the curriculum in Table \ref{tab:AAIresults}. This shows that after the same number of steps the VAE agent solves $93.6\%$ of the tasks and outperforms both IDF and RF with the RF agent significantly below the other two. This means that at the end of training when it is solving the most complex tasks the agent still maintains the ability to solve the easier tasks and has learned a more general solution.

\subsection{Test Score}

\begin{figure}[t]
\centering
\includegraphics[width =1\hsize]{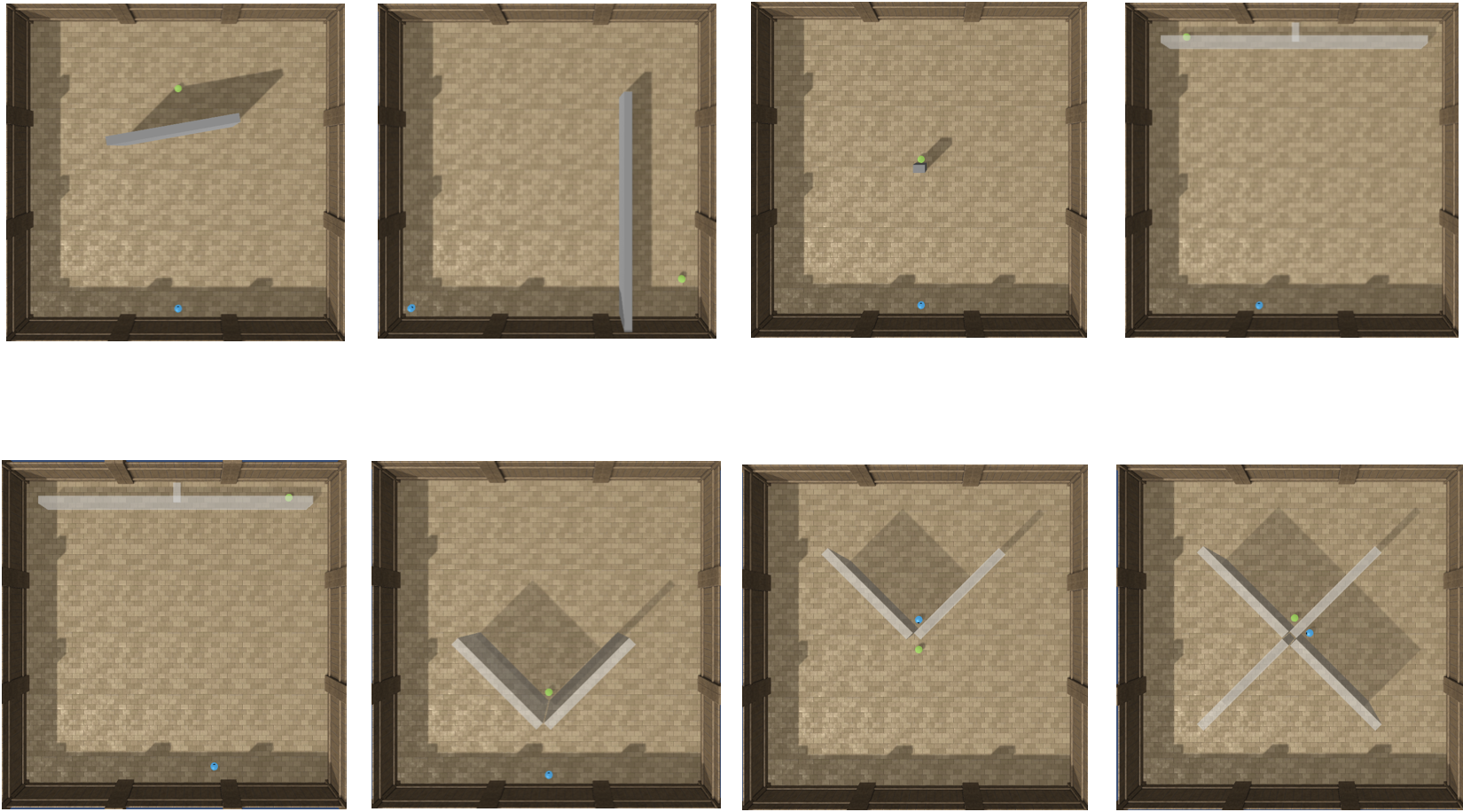}
\caption{The eight test arenas corresponding to 3-x-y in the AAIO testbed, where $x \in \{1,2,3,5,8,10,11,12\}$ and $y \in \{1,2,3\}$ }
\label{fig:test_curriculum}
\end{figure}

Finally, we examined the generalisation capabilities of the agents by testing them on the full set of detour tasks from the Animal-AI testbed as shown in Figure ~\ref{fig:test_curriculum}. We do not expect strong performances here as the test set includes object types unseen by the agents before, either in IDC or X-Maze curriculum or any pre-training of the VAE, such as opaque walls and half-height transparent walls.

Again, the VAE outperforms the two baselines, even solving some tasks in problems that contain objects that it has not encountered before. It also solves variations of the tasks with transparent walls that were unsolved in the Animal-AI competition, though this is not particularly surprising given that it was designed to solve them in particular whilst the competition entrants were unaware of the hidden X-Maze tasks and also trying to solve the rest of the large testbed. These generalisation results are promising for our approach. It is likely that a $\beta$-VAE trained on environments with opaque walls would score much better. Nevertheless, the main takeaway is the sample efficiency improvements over the training curriculum and the training scores.

\section{Comparison to Atari Breakout}\label{sec:resultsAtari}

\begin{figure*}[t]
    \begin{subfigure}{.5\textwidth}
        \centering
        \includegraphics[width =1\hsize]{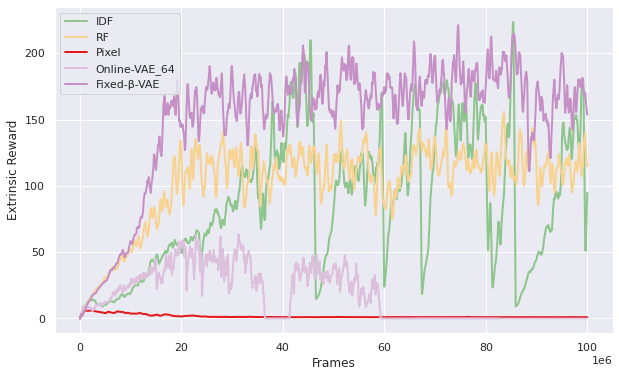}
        \caption{Extrinsic reward on Atari Breakout}
        \label{fig:breakout_extrinsic}
    \end{subfigure}
    \begin{subfigure}{.5\textwidth}
        \centering
        \includegraphics[width =1\hsize]{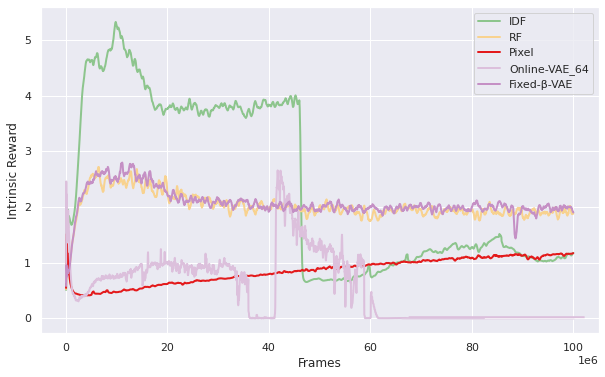}
        \caption{Intrinsic reward on Atari Breakout}
        \label{fig:breakout_intrinsic}
    \end{subfigure}
    \caption{Results comparing a fixed VAE encoding to previous work with online VAE, RF, IDF, and pixel encodings.}
    \label{fig:curriculum}
\end{figure*}

Given the success of using fixed $\beta$-VAEs in Animal-AI we decided to corroborate the results on a more commonly used benchmark. Due to limited computational resources we could not make a large-scale comparison so we limited to one environment. We chose Breakout for this as previous results showed a poor sample efficiency for the \emph{online} VAE compared to RF and IDF, requiring up to three times as much training steps to reach an equivalent score. As with previous experiments the agent learns purely from the curiosity reward signal and we ignore the extrinsic reward. Note that Atari is particularly well suited to such an approach because seeing new types of scene (and getting a high curiosity reward) is often correlated with progressing in the game (and getting a high extrinsic reward).

First, we attempted to replicate the results of Burda et al.\ with online VAE, IDF, RF, and raw pixel encodings. This was successful with the caveat that resource constraints restricted our experiments to 100M frames per agent. We managed to almost identically replicate the results for IDF, RF, and pixels, but received more unstable results for online VAEs as we had to scale down the number of simultaneous environments due to memory constraints. Nevertheless, the IDF, RF and pixel results provide a strong baseline for comparison to our approach which is repeated from the existing literature.

Fixed VAEs were previously not used before on Atari because, unlike Animal-AI, it is not well suited to generating a representative dataset for pre-training. This is because starting the environment and acting randomly leads to a very limited portion of the possible state space and one that is very different from the kind of states that a competent agent will encounter. There is also no clear method for randomly sampling from all possible game states. As we are not interested in the problem of solving the game here, and are simply performing additional experiments to corroborate our hypothesis, we create a dataset of observations from an already competent agent to create the VAE encoding. 

The results are shown in Figure \ref{fig:breakout_extrinsic} which compares the performance (extrinsic reward) of the five different agents. The fixed VAE agent outperforms the other encoding types. Note that the online VAE agent has very unstable performance. This can be understood by looking at a graph of the intrinsic reward over time as in Figure ~\ref{fig:breakout_intrinsic}. 

We observe a strong correlation of intrinsic reward and extrinsic performance, with the best curious agent being the ones with the most stable encoding network.
Furthermore, we observe that catastrophic forgetting happens (for IDF or VAE) when the forward prediction loss is volatile. This can come from an update of either the forward model or the encoding network, but given this happens only for unstable encoders is more likely the latter. Consequently, the encoder instability might be a strong source of catastrophic forgetting for curious agents.

We also notice that the fixed VAE and RF have very similar patterns for their intrinsic reward. This suggests that they are similar in terms of stability, so it is the content of the encoding of the VAE that is making the difference. Whilst we cannot conclude anything about solving Atari games from our results due to the use of Oracle observation data in the encoding, this does corroborate our previous results that, where they can be trained, the structure of VAEs can be leveraged by curious agents. 

\section{Conclusion and Future Work}\label{sec:conclusion}

One of the great challenges of reinforcement learning is to learn real-life tasks in environments with sparse rewards, fluctuating illumination, and 3D physics interactions between objects. Previous results had suggested that random features are sufficient for use with curiosity as they outperformed online VAEs on Atarti and similar benchmarks. In this paper we have shown the benefits of a factorised representation and how this can be used with curiosity. We tested fixed VAEs on the exploration intensive detour tasks from the Animal-AI Testbed. This environment makes it easy to pre-train a VAE from representative data prior to training an agent. Our methodology can be extended to any situation where pre-training of a VAE is possible. For example, in robotic grasping tasks, the $\beta$-VAE could be pre-trained with multiple observations taken from the environment \cite{curious-robot}.

We confirmed the benefits of using a fixed $\beta$-VAE encoding by analysing performance on a previously well-studied example problem. Here, the main problem is one of finding the data to train the VAE in the first place. We hope that RL research will move to more open-ended environments where such training data is more freely available and that this will open up research on structured representations with curiosity that can be used for domain general learning that can potentially even persist across multiple agents like the innate knowledge of some of the animals that can solve detour tasks at a young age. 

Nevertheless, for many existing environments it is not possible to pre-train a VAE that remains relevant as the agent explores. An obvious direction for future work is to explore methods to stabilise online VAE learning so that it can still be used with curiosity. For example, it may be possible to fix the VAE for extended periods or use constraints to ensure that it remains relatively stable as the agent's observations change. Details of the complex interplay between an agent's policy, its input encoding, and intrinsic reward, as they all evolve over time remains an interesting area for research.

Returning to the cognitive science motivations for using curiosity in AI, we can draw inspiration from the large amount of innate knowledge and structured encodings found in many animals. In the future we would like to try a hybrid approach exploring the interplay between strongly encoded innate knowledge and hyper-priors represented by the approach taken in this paper, combined with an evolving online VAE that updates to local changes in the incoming data distribution. The curiosity signal could be decomposed into a strongly weighted part from the fixed representation and a weakly weighted component form the online representation. It seems possible that the stability of the innate representation could be leveraged to keep the curiosity signal stable, whilst the online representation adapts to the specific task. Of course, much work remains to be done for this to work, but it could allow for domain general yet task specific intrinsic motivation to guide behaviour.

\newpage

\bibliographystyle{apalike}
\bibliography{reference}

\newpage

\section*{Appendix}

These supplementary materials give further details on the experimental design and results. All the simulations were computated on two NVIDIA Titan Z GPUs with 32GB of RAM.  

\subsection*{Animal-AI}

We include first further experimental details and results regarding the pre-training on the Initial Detour Curriculum (IDC). We also include further results and details about the X-Maxe Curriculum (XMC) training. 

\subsubsection*{IDC preliminary experiments}

\begin{figure}[h]
\centering
\includegraphics[width=0.4\textwidth]{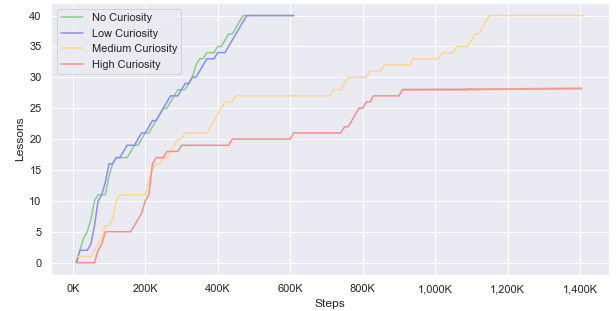}
\caption{Lesson Evaluation on the IDC.}
\label{fig:exp1}
\end{figure}

Figure~\ref{fig:exp1} shows the effect of curiosity on the IDC. The IDC is used so that the agent has the basic navigational abilities necessary for performance in the full XMC experiments. Without this, the XMC results would also include variation for learning non-detour type task skills, which would detract from our exploration of curiosity in detour tasks.

In this first experiment, four agents with different curiosity weight $\lambda_c$ are compared.
The green ($\lambda_c=0$) and blue ($\lambda_c=0.01$) agents can complete the IDC under half a million steps, the high curious agents are less sample efficient: the yellow  ($\lambda_c=0.1$) agent takes more than a million steps and the red  ($\lambda_c=0.5$) got stuck at lesson 28. 

\begin{figure}[H]
\centering
\includegraphics[width = 0.8\hsize]{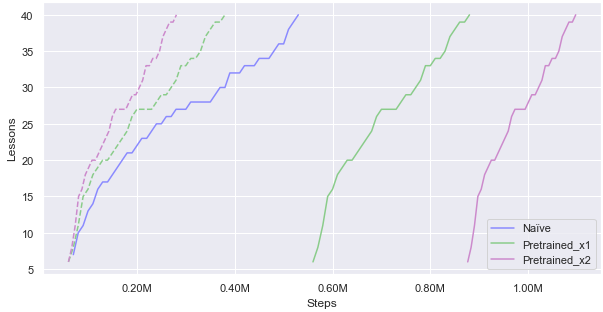}
\caption{Lesson evolution of a low curious agent during three runs on the IDC. The dotted lines show the second and third runs transposed to overlay the initial run to aid visual comparison.}
\label{fig:exp2_a}
\end{figure}

\begin{figure}[H]
\centering
\includegraphics[width = 0.8\hsize]{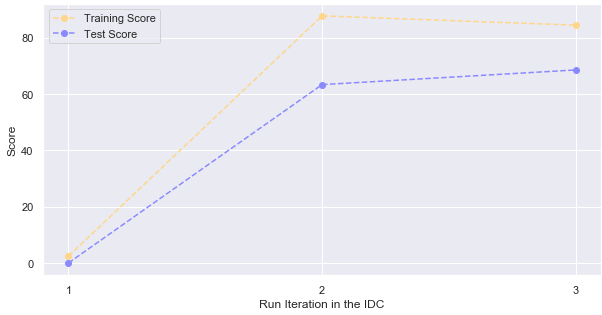}
\caption{Scores of the low curious agent after each of the three IDC runs.}
\label{fig:exp2_b}
\end{figure}

Figures~\ref{fig:exp2_a} and~\ref{fig:exp2_b} show the results from cycling through the curriculum. First, we observe that agents manage to complete the curriculum faster after each iteration, with roughly 30\% gain in sample-efficiency. The second and third runs curves are replicated in dotted lines along the first run curve. In under 1.2 million steps, the agent is able to perform three runs of the IDC. From Figure \ref{fig:exp2_b}, we observe that agents need more than one run on the IDC to learn the targeted abilities robustly. After a single run the agent only managed a training score of 2\%, having `unlearned' to solve the majority of tasks. By the end of the third run it manages to reach 82\% on the training set and also can solve 67\% of unseen, though similar, environments.

\begin{figure}[H]
\centering
\includegraphics[width = .8\hsize]{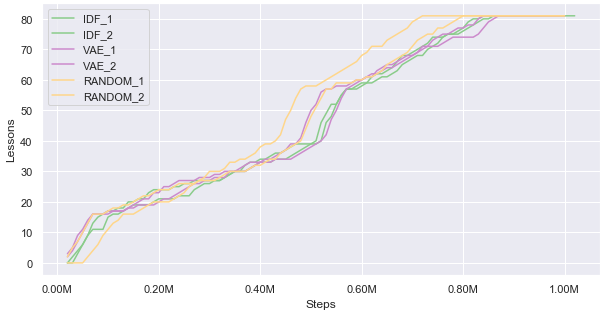}
\caption{Lesson Evolution of the six agents during the IDC pre-training.}
\label{fig:exp5}
\end{figure}

For the full experiments we started with weights from an non-curious agent that had been trained for 1 million steps on the IDC. Figures \ref{fig:exp5} shows the results from training 6 different agents and it can be seen that they follow roughly similar learning curves as at this stage (without curiosity) they behave identically.

\subsubsection*{XMC preliminary experiments}

\begin{figure}[H]
\centering
\includegraphics[width = 0.8\hsize]{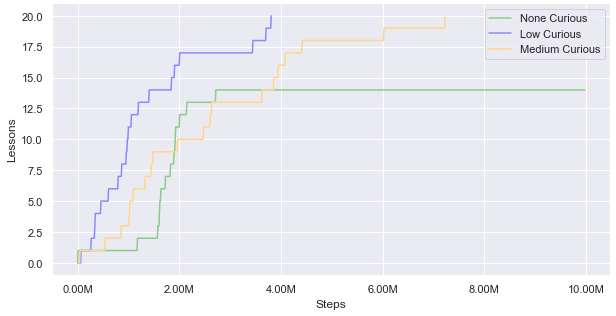}
\caption{Lesson evolution of  curious agents on the XMC.}
\label{fig:exp3}
\end{figure}

Figure~\ref{fig:exp3} shows the effect of the curiosity parameter for learning in the XMC (without the initial pre-training on IDC outlined above). A non-curious agent is unable to complete the XMC curriculum. Moreover, we observe that the low curious agent, with a curiosity weight of 0.01, is the most sample efficient. Note that these preliminary experiments used a slightly different version of XMC than the final version presented in the paper.

\begin{figure}[H]
\centering
\includegraphics[width = 0.8\hsize]{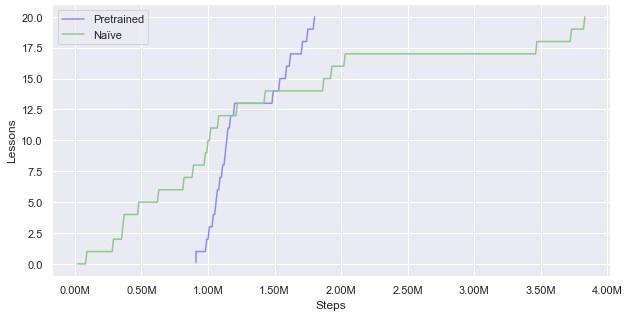}
\caption{Lesson evolution of two low curious agents on the IDC.}
\label{fig:exp4}
\end{figure}

Figure~\ref{fig:exp4}, shows the impact of pre-training a low curious agent on the IDC (two runs) before training it on the XMC. The pre-trained agent took less than a million steps to complete the XMC while the naive agent took almost four million steps. Even if we include the two runs on the IDC, we can see that the pre-training divided the number of steps by a factor of two.

\subsubsection*{Extra $\beta$-VAE Network details }

\paragraph{Network Architecture}

\begin{figure}[h]
    \centering
    \includegraphics[width=0.4\textwidth]{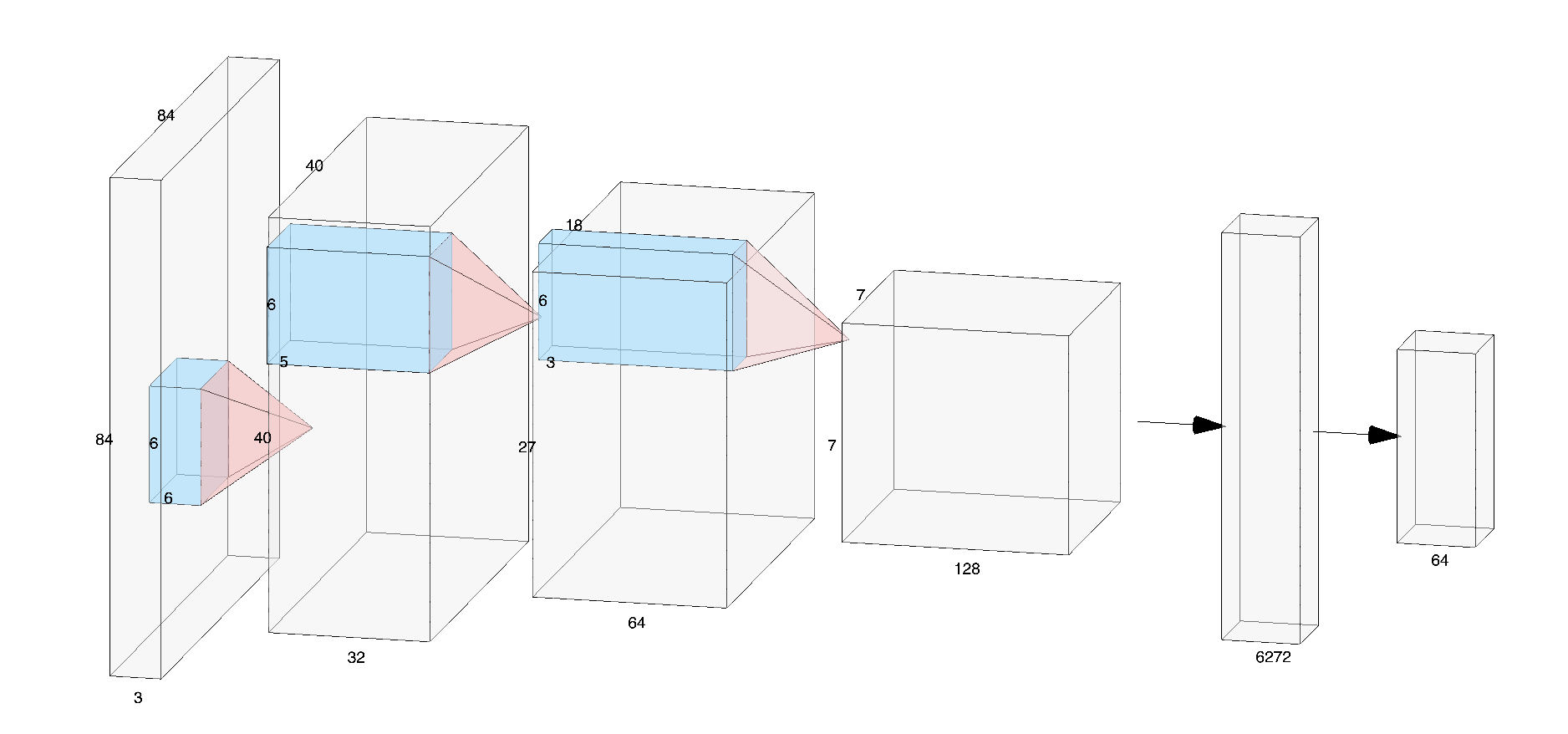}
    \caption{Architecture of the VAE encoder.}
    \label{fig:vae-architecture-encoder}
\end{figure}

\begin{figure}[h]
    \centering
    \includegraphics[width=0.4\textwidth]{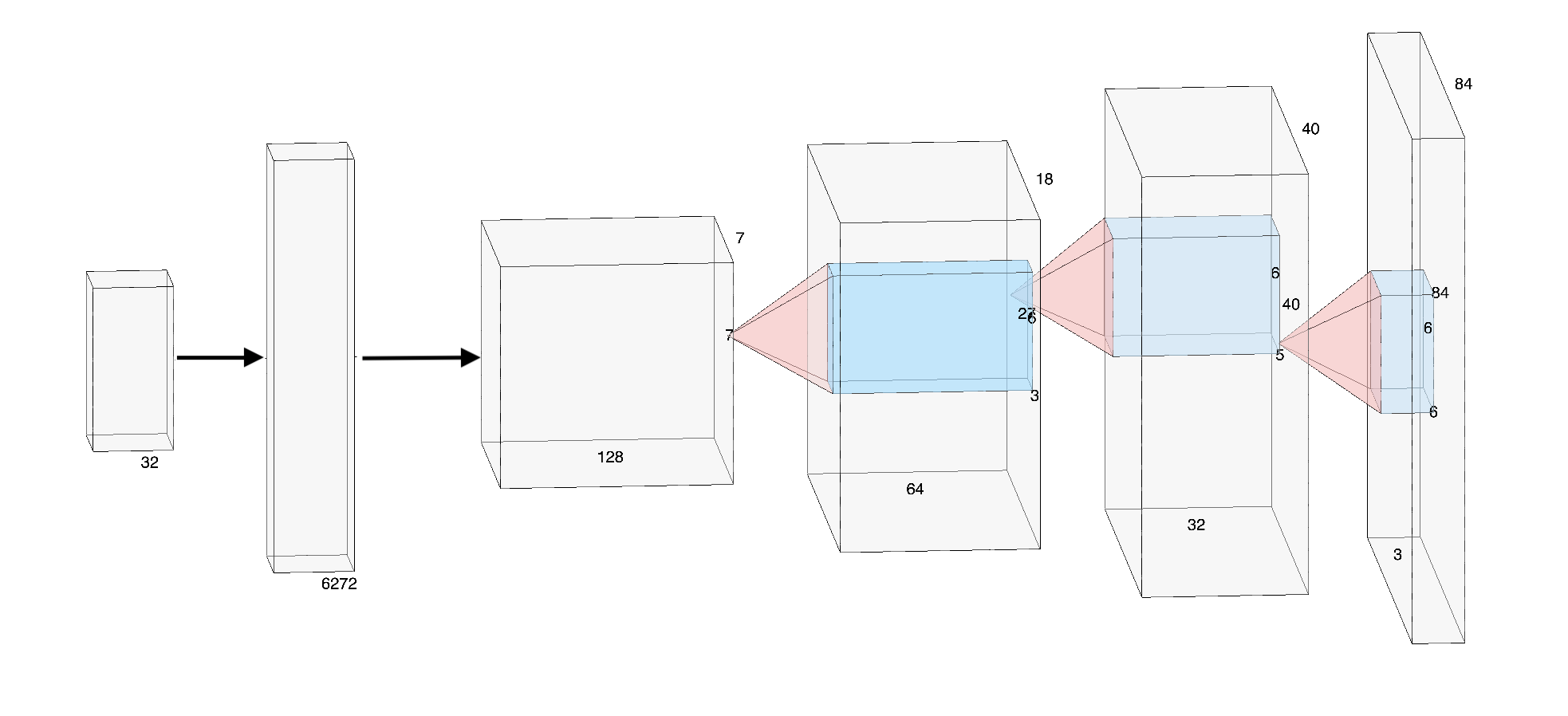}
    \caption{Architecture of the VAE decoder.}
    \label{fig:vae-architecture-decoder}
\end{figure}

Figures~\ref{fig:vae-architecture-encoder} and \ref{fig:vae-architecture-decoder} illustrate the overall network structures used for the encoder  and the decoder respectively.

\subsubsection*{Final Experiment Supplementary Details}

For the final experiments agents were first trained on the IDC for 1 Millions steps to obtained the required basic abilities. Then, they were trained on the XMC for 2.5 Millions steps. From each of the pre-trained agents (see Figure \ref{fig:exp5}) we were able to train five resulting agents. This gives us ten trained agents for each configuration, amounting to 30 agents in total. 

\begin{figure}[H]
\centering
\includegraphics[width = 1\hsize]{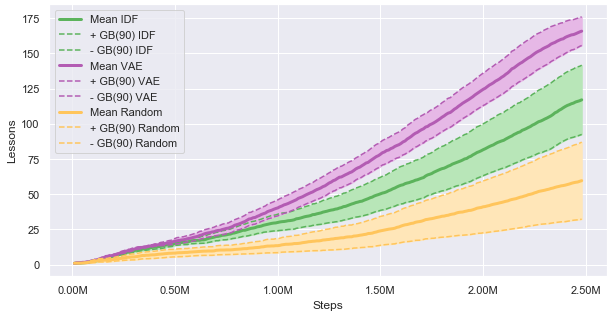}
\caption{Mean Lesson Evolution of agents during the XMC training.}
\label{fig:exp6_mean}
\end{figure}

\begin{figure}[H]
\centering
\includegraphics[width = 1\hsize]{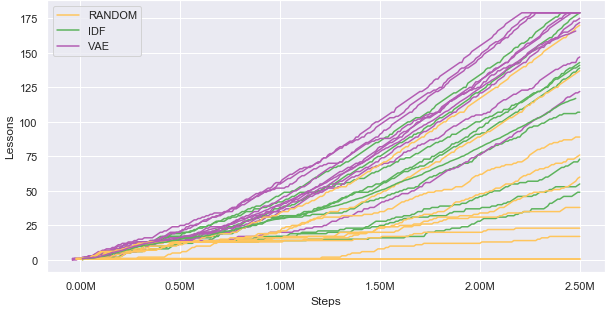}
\caption{Lesson Evolution of the 30 agents during the XMC final training. }
\label{fig:exp6_all_runs_pr}
\end{figure}

\bigskip
Figure~\ref{fig:exp6_mean} is our main results (also in the main paper). Note that the displayed confidence intervals rely on the assumption that the results follow Gaussian distributions. Figure \ref{fig:exp6_all_runs_pr} shows all the runs on the same plot.

\subsection*{Atari Breakout Environments}

\subsubsection*{Reproducing Previous Results}

The results were obtained from a fork of the Large-Scale Curiosity paper github repository with default hyperparamaters. Due to computational resource constraints we limited the number of frames (training steps) to 100 Million, six times less than in the original paper. For the online VAE training, we were able to use only 64 environments in parallel due to memory constraints. The reproduction results are illustrated in the paper and match rather closely the results claimed in the Burda paper.

\subsubsection*{Extra Fixed $\beta$-VAE Training details}

\paragraph{Data Collection}

The method we propose requires to pre-train a $\beta$-VAE offline with collected observations. These observations distribution should match closely the distribution of observations the subsequent agent will encounter. As we can't perform random spawning like in Animal-AI, we need to rely on an already trained agent to collect the data. The dataset is comprised of 100K observations collected during the last iterations of the IDF agent. We obtain an observation distribution of states with extrinsic reward ranging from 0 to 200 with a higher density for small reward states.

\paragraph{Training $\beta$-VAE}
The $\beta$-VAE was trained on the dataset for 5 epochs. We used a latent space dimension of 32 and a $\beta$ regularisation of 10---the same as in the Animal-AI experiments. We obtained a small reconstruction loss as only few pixels are changing from one frame to another.

\end{document}